\pdfoutput=1

\documentclass[11pt]{article}

\usepackage{acl}

\usepackage{times}
\usepackage{latexsym}

\usepackage[T1]{fontenc}

\usepackage[utf8]{inputenc}

\usepackage{microtype}

\usepackage{algorithm}
\usepackage{algorithmic}
\usepackage{graphicx}
\usepackage{caption}
\usepackage{subcaption}
\usepackage{diagbox}
\usepackage{expex}
\usepackage{dialogue}
\usepackage{microtype}
\usepackage{makecell}
\usepackage{booktabs}
\usepackage{todonotes}
\usepackage{adjustbox}
\usepackage{xcolor}
\definecolor{darkgreen}{RGB}{0,100,0}

\title{When to generate hedges in peer-tutoring interactions}
\author{Alafate Abulimiti$^{1,2}$,
Chloé Clavel$^{3}$, 
Justine Cassell$^{1,4}$
 \\~\\
 $^1$  INRIA, Paris $^2$ ENS/PSL 
  \texttt{<alafate.abulimiti@inria.fr>} 
\\
$^3$ LTCI, Institut Polytechnique de Paris, Telecom Paris 
{\tt <chloe.clavel@telecom-paris.fr>}\\
$^4$  Carnegie Mellon University 
\texttt{<justine@cs.cmu.edu>}
}

\begin{document}
\maketitle
\begin{abstract}

This paper explores the application of machine learning techniques to predict where hedging occurs in peer-tutoring interactions. The study uses a naturalistic face-to-face dataset annotated for natural language turns, conversational strategies, tutoring strategies, and nonverbal behaviors. These elements are processed into a vector representation of the previous turns, which serves as input to several machine learning models, including MLP and LSTM. The results show that embedding layers, capturing the semantic information of the previous turns, significantly improves the model's performance. Additionally, the study provides insights into the importance of various features, such as interpersonal rapport and nonverbal behaviors, in predicting hedges by using Shapley values \citep{hart1989shapley} for feature explanation. We discover that the eye gaze of both the tutor and the tutee has a significant impact on hedge prediction. We further validate this observation through a follow-up ablation study.

\end{abstract}

\section{Introduction}

Effective communication involves various conversational strategies that help speakers convey their intended meaning and manage social interactions at the same time. These strategies can include the use of self-disclosure, praise, reference to shared experience, etc. \citep{zhao2014towards}. Hedges are one of those strategies that is commonly used in dialogue. Hedges are words or phrases that convey a degree of uncertainty or vagueness, allowing speakers to soften the impact of their statements and convey humility or modesty, or avoid face threat. Although hedges can be effective in certain situations, understanding when and how to use hedges is essential and challenging.

The use of hedges is especially significant in tutoring interactions where they may facilitate correcting a wrong answer without embarrassing the recipient. However, the use of hedges in this context is not limited to expert educators. They are also found to be abundant in peer-tutoring settings. In fact,
\citep{madaio2017impact} found that confident tutors tend to use more hedges when their rapport with the tutee is low, and that this pattern leads to tutees attempting more problems and solving more problems correctly. Hence, the detection and correct deployment of hedges, at the right time, is not just pleasant, but crucial for the development of effective intelligent peer tutoring systems.

While the use of hedges in conversation is an important aspect of effective communication, automatically generating hedges in real-time at the right time, can be a challenging task. In recent years, there have been several studies of automatic hedge detection \citep{raphalen-etal-2022-might, goel2019a}, particularly in the context of dialogue systems. However, despite significant advances in detection, generating hedges in a timely and appropriate manner remained unsolved. For example, the RLHF-based training method enables the development of robust language models that align with human preferences \citep{ouyang2022training}. However, this approach does not explicitly instruct large language models (e.g., ChatGPT) in pragmatic and social skills, such as the appropriate use of hedges during communication. This lack of specific training can result in a gap in the model's ability to effectively integrate these conversational nuances into its responses in \textit{at the correct time}. This limitation can affect the quality of communication and highlights the need for further research on effective hedge strategie generation; that is, to generate hedges at the right time.

Despite the widespread use of hedges in communication, there is still much to learn about their timing and the effectiveness of their use, particularly in dialogue rather than running text, and specifically in the current article, in peer-tutoring environments.

To address this gap in the literature, our research focuses on two key questions:

\textbf{RQ1}: First, can we predict when hedges should be generated in peer-tutoring environments? \label{rq1}

To address this question we investigate whether it is possible to identify the points at which hedges should be introduced during a peer tutoring dialogue.

\textbf{RQ2}: Second, what features contribute to accurate predictions? of where to place hedges? \label{rq2}

To address this question we focus on the explainability of classification models using Shapley values \citep{sundararajan2020many} .

\section{Related Work}

\subsection{Hedges}
Hedges are a common rhetorical device used to diminish the impact of an utterance, often to avoid unnecessary embarrassment on the part of the listener or to avoid the speaker being interpreted as rude. In linguistic terms, hedges diminish the full semantic value of an expression \citep{fraser2010a}. \textbf{Propositional hedges}, also called \textit{Approximators}, refers to the use of uncertainty \citep{vincze2014uncertainty}, vagueness \citep{williamson2002vagueness}, or fuzzy language \citep{lakoff1975hedges}, such as ``sort of'' or ``approximately''. On the other hand, \textbf{Relational hedges} are used to convey the subjective or opinionated nature of a statement, such as ``\textit{I guess} it will be raining tomorrow.''. \textbf{Apologizer} \citep{raphalen-etal-2022-might, goel2019a, fraser2010a}, which is an expression used to mitigate the strength of an utterance by using apologies, is another type of hedges. such as ``\textit{I am sorry}, but you shouldn't do that.'' Although the different types of hedges function differently, they all share a common role of mitigation in conversation. Therefore, in this paper, we focus on simply predicting hedges vs non-hedges. 

As described above, in tutoring, including peer tutoring, hedges are frequently used and have a positive impact on performance \citep{madaio2017impact}. Powerful language models such as GPT-4 \citep{GPT42023} and ChatGPT \citep{chatGPT2023} are now capable of generating hedges with appropriate prompts, but these language models do not actively generate hedges \citep{abulimiti2023}, fIn other words, the question of how to use thedges correctly in the next conversational action remains unsolved.

\subsection{Conversational Strategy Prediction}
The development of approaches for predicting conversational strategies -- or particular ways of saying things -- has progressed significantly over the past few years in the field of dialogue systems. Early studies, such as the COBBER, a domain-independent framework, used a Conversational Case-Based Reasoning (CCBR) framework based on reusable ontologies \citep{gomez2006conversational}. The aim was to help people use a computer more effectively by keeping them in the right mood or frame of mind. Methods such as reinforcement learning have also been introduced in non-task-oriented dialog systems, including a technique known as policy learning \citep{yu2016strategy}.   Reinforcement learning has been explored, as well,for training socially interactive agents that maximize user engagement \citep{galland2022adapting}. 

The Sentiment Look-ahead method is used to predict users' future emotional states and to reward generative models that enhance user sentiment\citep{shin2020generating}. The rewards include response relevance, fluency, and emotion matching. These rewards are built using a reinforcement learning framework, where the model learns to predict the user's future emotional state. \citet{romero2017cognitive} designed a social reasoner that can manage the rapport between user and system by reasoning and applying different conversational strategies. 

More recently, deep learning-based approaches have emerged. For example, the Estimation-Action-Reflection (EAR) framework combines conversational and recommender approaches by learning a dialogue policy based on user preferences and conversation history \citep{lei2020estimation}.

Perhaps the most recent advances in the field have focused on how to create an empathetic dialogue system. MIME \citep{majumder2020mime} used the emotion mimicry strategy to match the user's emotion based on the text context. EmpDG \citep{li2020empdg} generated empathetic responses using an interactive adversarial learning method to identify whether the responses evoke emotional perceptivity (the ability to perceive, understand, and be sensitive to the emotions of others. ) in dialogue.  The Mixture of Empathetic Listeners (MoEL) model \citet{lin2019moel} generates empathetic responses by recognizing the user's emotional state, using emotion-specific multi-agent listeners to respond, and then combining these responses based on the emotion distribution. This process effectively merges the output states of the listeners to create an appropriate empathetic response. The model then crafts an empathetic response grounded in the user's emotions, which are monitored by the emotion tracker. Despite the notable success of MIME and MoEL in predicting emotions or conversational strategies, they do not incorporate the social context (e.g., the relationship between speakers), or the emotional tenor of the conversation up until that point, nor do they include important nonverbal behaviors into reasoning and decision-making processes.  However, such elements are fundamental for the correct use of social language, and their absence potentially limits the effectiveness and naturalness of these models. 

Predicting the appropriate emotion or conversational strategies in a conversation is a challenging task, mainly because determining what is ``appropriate'' in a conversation is rather subjective and is certainly context-dependent. For example, EmpDG \citep{li2020empdg} model achieved an accuracy of approximately 0.34 across the 32 evenly distributed labels in the Empathetic Dialogue dataset \citep{rashkin2019towards}. indicating the complexity of the problem at hand. Similarly, MoEL \citep{lin2019moel} model achieved varying degrees of accuracy in the same dataset - 38\% for the top 1, 63\% for the top 3, and 74\% for the top 5 for emotion detection, further emphasizing the difficulty of the task. 

The current paper aims to fill the lucunae in prior work by integrating social context and nonverbal behaviors as predictive features to construct predictive models for hedges.  


\section{Methodology}

\subsection{Task Description}
Suppose we have a set of dialogues $D=\{d_1,d_2,d_3,...d_n\}$. Each dialogue $d=\{u_1, u_2, u_3...u_m\}$ consists of $m$ turns, with $u_i$ representing a specific turn. Both tutor and tutee turns in these dialogues can be categorized as either hedges or non-hedges. However, for the purposes of our analysis, we will primarily focus on the tutor's turns. The label of a particular turn $u_i$ is denoted as $l_i$. Furthermore, every turn can be depicted as a feature vector $X$, composed of elements $(x_1, x_2, ..., x_N)$. Here, $N$ signifies the total number of features used to characterize a turn. Each turn in the dialogue is assigned a fixed window size ($\omega$) of the dialogue history, represented as: $h_i=\{u_{max(1, i-\omega)}, u_{i-\omega+1},...u_{i}\}$. The primary objective of this research is to develop a model, denoted $M$, capable of predicting the type of hedge $l'_{i+1}$ that a tutor will use next, based on the dialogue history $h_i$. The effectiveness of the model is measured using standard classification metrics, such as precision, recall, and  F1 score.

Predicting hedges in a peer-tutoring conversation can be simplified to a binary classification problem. The features used as inputs are extracted from the turns in the interaction (further details in Section \ref{subsec:features}), while the output is a binary value showing whether or not hedges are present in each turn. 

\subsection{Corpus}

The dataset used in the current work is the same as that employed in our previous work on hedges  \citep{madaio2018climate}. It is a subset of a larger investigation into the role of social, rapport-building conversational strategies in task-oriented dialogue. The corpus consists of face-to-face interaction from 20 same-gender dyads of American teenagers, with an average age of 14.3 years (and a range of ages from 13 to 16 years), gender-balanced \footnote{The corpus used here comes from earlier work by
the last author and her colleagues, as cited above, and was used in accordance with the original experimenters’ Institutional Review Board (IRB) approval. That approval required that the children's data not be released, which means that we cannot share the corpus. However, a pixelated example of the video data is available at \url{github.com/neuromaancer/hedge_prediction}.} , and recorded twice over two weeks. However, due to technical issues, data from only 14 dyads' data were usable. The participants were asked to to take turns tutoring one another in different aspects of linear algebra. Each hour-long session was divided into 4 phases: an initial social period, followed by a first peer tuturing period, then a second short social period, and finally, the teens switched roles, with the tutee becoming tutor for the second task period.  For the 14 dyads we used for our model, 28-hour-long face-to-face interactions were recorded over the period of two weeks.  The recorded video and audio data were transcribed, resulting in approximately 9479 turns for the 14 dyads. These included 8399 non-hedges and 1080 hedges. 4214 non-hedges and 507 hedges in the tutors' turns since, as described above, we looked only at tutor hedges for this analysis, (although note that both tutor and tutee hedges in prior turns were used as input)). A ''hedge turn'' is any turn that includes hedging language. We also retained non-speech segments such as laughter and fillers. 

Peer tutoring is a popular teaching method used in many schools and educational settings. As described above, and in  \citep{madaio2017think}, even though these teenagers may be inexperienced, in contexts of low rapport, when they use hedges during tutoring, their tutees are encouraged to attempt more problems and succeed in solving more of them. This positive outcome justifies the use of this dataset for studying hedges in tutoring interactions. While we recognize the importance of exploring the use of hedge with expert tutors in the future, our current focus on untrained peer tutors provides a unique perspective on how hedges can impact learning, even when the tutors themselves are not highly experienced. The methods and results from our study can be used as a foundation for future research, which could include the investigation of expert tutors and the potential differences in their use of hedges.

\subsection{Features}\label{subsec:features}
In this section, we outline the features used as input vectors (i.e., $u_i$ vector) for our prediction model, which seeks to properly predict the hedging strategy for the tutor's upcoming turn. In total, we have a vector with a length of 438 to represent a turn. 

\subsubsection{Turn embedding}
Turn embedding is a common technique in natural language processing that involves representing a turn as a vector. In this study, we apply a sentence transformer \citep{reimers2019sentence} to generate turn embeddings from the tutor-tutee conversation. This feature enables us to capture the semantic meaning of the turn in the context of the conversation, which can be helpful for predicting hedges. 

\subsubsection{Conversational Strategies (\textit{CS}) of the previous turns}

Conversational strategies refer to the different ways of speaking used by both speakers to manage social interaction. Strategies considered in this study are self-disclosure, praise, violation of social norms, and hedges. Self-disclosure \citep{derlega1993self} refers to situations in which the tutor or tutee shares personal information, which is often used to build rapport. Praise \citep{brophy1981teacher} is a form of positive feedback that acknowledges and reinforces the other person's behaviors or attributes. Violation of social norms \citep{zhao2014towards}, which in this population often consists of friendly teasing, is a conversational move in speaker demonstrates the special nature of the relationship with the listener by engaging in slightly transgressive behavior. The conversational strategy annotation was carried out by \citet{madaio2018climate}, and inter-rater reliability achieved a minimum Krippendorff's alpha of over .7 for all strategies.

In terms of hedges, we note that we only use the speakers' previous hedge strategies to predict the  tutor's next hedge strategy. This avoids any issues with predicting label leakage. 

\subsubsection{Tutoring Strategies (\textit{TS}) in the previous turns}
Tutoring strategies \citep{madaio2016effect} refer to the different techniques employed by the tutor or tutee to facilitate learning. Strategies considered in this study include deep / shallow questions, meta-communication, knowledge building, and knowledge telling. The deep question encourages critical thinking and higher-order cognition. The shallow question is used to confirm or clarify understanding. Meta-communication is a strategy whereby the tutor or tutee refers to the tutoring process or the tutor / tutee's self-evaluation of their own knowledge, which can help to clarify misunderstandings and promote effective communication. Knowledge building involves introducing new concepts or ideas, discussing the reasoning-mathematical solving steps, and providing examples. Knowledge telling refers to providing information (i.e., simply stating numbers, variables).  The tutoring strategies annotation was also carried out by \citet{madaio2018climate}, with annotators achieving a minimum Krippendorff's alpha of .7 for all tutoring strategies.

\subsubsection{Dialogue Act (\textit{DialAct}) of the previous turns}
Dialogue acts are types of speech acts \citep{searle1965speech} used by tutors and tutees during their interactions. In our study, we use the widely-used DAMSL (Dialogue Act Markup in Several Layers) \citep{jurafsky1997switchboard} coding schema to annotate dialogue turns by using a state-of-the-art dialogue act classifier with context-aware self-attention \citep{raheja2019dialogue}. In our dataset, only 6 dialogue acts were found, they are \textit{Abandoned} or Turn-Exit (\textit{\%}) , \textit{Acknowledge (Backchannel)} (\textit{b}), \textit{Backchannel in question form} (\textit{bh}), \textit{Yes-No-Question} (\textit{qy}), \textit{Statement-non-opinion} (\textit{sv}) and \textit{Statement-opinion} (\textit{sd}).

\subsubsection{Rapport in the previous turns}

As our previous work demonstrates, the level of rapport between tutor and tutee plays a role in the use of hedges. We therefore include it as a feature in our study.
Rapport is ``The relative harmony of relations felt by both participants'' \citep{spencer2005politeness}. The rapport annotation was carried out by Amazon Mechanical Turk (AMT) annotators as described in \citep{madaio2018climate}. Rapport level was operationalized as a $7$ point Likert scale, where a higher score indicates a stronger level of rapport. For the annotation of rapport, the annotators employed the ``thin slice'' method \citep{ambady1993half}, whereby the experimenter segmented each video into 30-second clips and randomized the order. To ensure the quality of rapport annotations, three annators evaluated each clip, and the experimenter applied the inverse-bias correction method \citep{parde2017finding} for selecting a single score for each clip. In the current study, when the dialogue history is contained within a single slice, we directly use the annotated rapport level of that particular slice as the historical rapport level. However, if the dialogue history extends over two slices, we select the rapport level of the slice containing the majority of the dialogue history.

\begin{figure*}
    \centering   \includegraphics[width=0.7\linewidth]{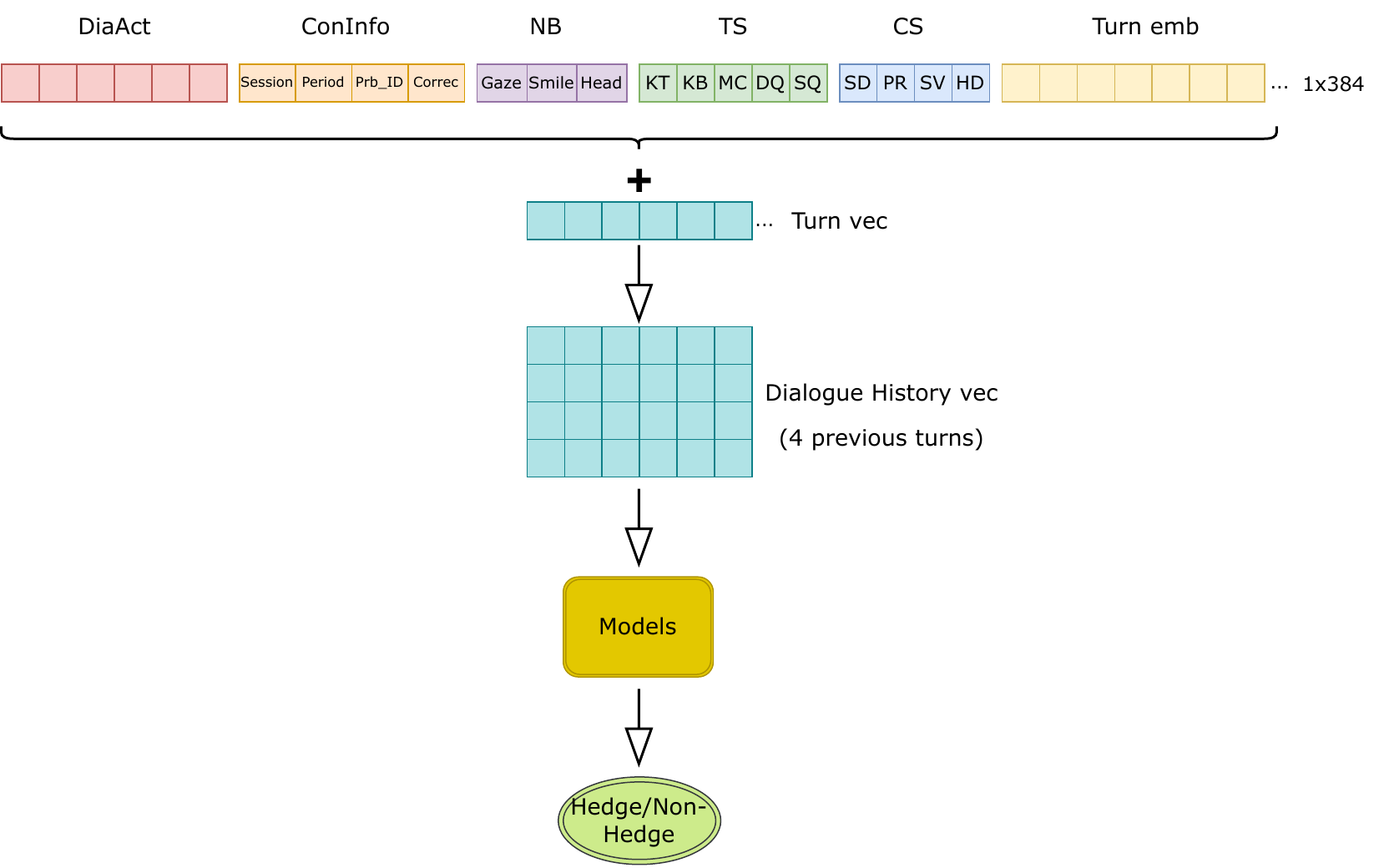}
    \caption{Vector Representation}
    \label{fig:vec}
\end{figure*}

\subsubsection{Nonverbal Behaviors (\textbf{NB})}
 Nonverbal behaviors, such as head nod, smile, and gaze, are an essential aspect of interpersonal communication that can also contribute to the development of rapport \citep{tickle1990nature}. The gaze and smile annotation was carried out by \citet{madaio2018climate}, we  annotated the head nods with 2 annotators. All the annotations were carried out after annotators reached an inter-rater reliability of  $0.7$ or above on Krippendorff's alpha. We collected all nonverbal behaviors that occurred during one turn and encoded them using one-hot encoding. For head nods and smiles, we used a binary labeling approach, marking $1$ for their occurrence and $0$ for non-occurrence. As gaze serves as a potent indicator of attention, we categorized it into 4 distinct types: no gaze appeared in the video, gaze at partner, gaze at worksheet, and gaze elsewhere. 

Mutual gaze between interlocutors, mutual smiles, and mutual head nods serve as great indicators of alignment and rapport in communication. These are not encoded separately, as our encoding process for nonverbal behaviors captures the behaviors of both participants within a turn, not only the current turn holder.'s Our current approach successfully captures these important  mutual signals.

\subsubsection{Contextual Information (\textit{ConInfo}) in the previous turns}

Our model also incorporates contextual information that characterizes the discourse environment between the two interlocutors. Specifically, we include features such as the session and period numbers, which help to encapsulate the temporal dynamics of the tutoring interactions. We also consider the math problem ID and the correctness of the current problem response, which act as markers of the present learning context. These features can illuminate the complexity of the ongoing problem and the students' performance, potentially influencing their use of hedges. The tutee's and tutor's pre-experiment test scores are also included, serving as initial measures of their knowledge before the tutoring session. This data can help to identify the starting knowledge disparity between the tutor and the tutee. It is plausible that these pre-test scores might also be linked with the students' level of confidence, which could subsequently impact their use of hedges. \citep{madaio2017impact}.

\citet{norman2022studying} suggests a link between verbal alignment signals, such as backchannels (e.g., ``um'', ``hhm'', ``oh..''), and learning gains in a cooperative learning environment. Given the role of hedging as a social language skill that improves learning performance, we hypothesize its connection to dynamic learning gains. Consequently, we incorporated the frequency of these verbal alignment signals from the previous four conversational turns into our model input.

\subsection{Vector Representation}
Before presenting the specific models, we first describe how we convert each sequence of turns into a vector representation. Our vector representation consists of three basic parts: turns as a sequence of tokens, annotations based on the turn (e.g., conversational strategies), and the nonverbal behaviors. Figure \ref{fig:vec} shows that we divide a vector of turns into 6 parts: turn embedding, conversational strategies (\textit{CS}), tutoring strategies (\textit{TS}), nonverbal behaviors (\textit{NB}), contextual information (\textit{ConInfo}) and dialogue acts (\textit{DialAct}). After encoding each turn in this fashion, we use the four previous turns as a history tensor of a turn. This history ten tensor will be the input to the prediction models, and the model's output will be this turn's hedge label.

\subsection{Prediction as Classification}
We mentioned in the previous section that we transform the prediction problem into a classification problem. This means that the corresponding hedge strategy is obtained by classifying different previous interactions (i.e., dialogue history) and historical characteristics (e.g., rapport, etc.). The classification models used are presented here. 

The selection of learning models in this study is strategic and based on our research objectives. Our primary aim is not to engineer a perfect system for hedging. Instead, we seek to comprehend the variables that influence hedging in dialogue. As such, our approach leans towards the use of models that are effective in contextual understanding. For example, Long Short-Term Memory networks (LSTMs) were chosen over Multi-Layer Perceptrons (MLPs) due to their superior ability to manage and interpret context, an essential factor in our exploration of hedging phenomena.

\subsubsection{LightGBM}
In this work, we used LightGBM \citep{ke2017lightgbm}, a gradient boosting framework known for its efficiency. We use it to predict hedges in dialogues, relying only on dialogue features such as conversational strategies, tutoring strategies, nonverbal behaviors, and contextual information, while turn embeddings are not included.

\subsubsection{XGBoost}
We also used the Extreme Gradient Boosting (XGBoost) algorithm \citep{chen2016xgboost}, which is a decision tree-based ensemble machine learning algorithm that uses a gradient boosting framework. Similar to LightGBM, the turn embedding is not used.

\subsubsection{Multi-layer perceptron (MLP)}
We constructed a multi-layer perceptron using two sets of features. These included a pre-trained contextual representation of the turn, specifically from the SentBERT model \citep{reimers2019sentence} which is the most prevalent sentence embedding tool, and the concatenation of all the features mentioned in Section \ref{subsec:features}.

\subsubsection{Long Short-Term Memory (LSTM)}
We use the same features and apply them to LSTM \citep{hochreiter1997long} and also LSTM with attention \citep{bahdanau2015neural}. LSTM has a good ability to capture temporal correlations, and we expect this ability to enhance prediction performance.

\begin{table*}[htbp]
  \centering
  \resizebox{0.7\textwidth}{!}{%
    \begin{tabular}{lccc}
    \toprule
    \textbf{Models} & \textbf{F1-score} & \textbf{Precision} & \textbf{Recall} \\
    \midrule
    LightGBM (w/o emb)  & 0.24 ($\pm 0.07$)   & 0.17 ($\pm 0.03$)    & 0.45 ($\pm 0.07$) \\
    XGBoost (w/o emb)  & 0.24 ($\pm 0.07$)   & 0.16 ($\pm 0.03$)    & 0.45 ($\pm 0.07$) \\
    \midrule
    MLP   & 0.25 ($\pm 0.06$)   & 0.16 ($\pm 0.03$)  & 0.52 ($\pm 0.07$) \\
    MLP (only emb) & 0.26 ($\pm 0.05$) & 0.16 ($\pm 0.02$) & 0.74 ($\pm 0.06$) \\
    MLP (w/o emb) & 0.26 ($\pm 0.06$) & 0.17 ($\pm 0.06$) & 0.56 ($\pm 0.07$)  \\
    \midrule
    LSTM  & 0.25 ($\pm 0.06$)   & 0.16 ($\pm 0.03$)  & 0.50 ($\pm 0.07$) \\
    LSTM (only emb) & 0.28 ($\pm 0.07$)  & 0.19 ($\pm 0.08$) & 0.52 ($\pm 0.07$) \\
    LSTM (w/o emb) & 0.25 ($\pm 0.05$)  & 0.15 ($\pm 0.02$) & 0.75 ($\pm 0.06$) \\
    AttnLSTM & 0.24 ($\pm 0.06$) & 0.15 ($\pm 0.03$) & 0.57 ($\pm 0.07$) \\
    AttnLSTM (only emb) & 0.25 ($\pm 0.07$) & 0.17 ($\pm 0.03$) & 0.45 ($\pm 0.07$) \\
    AttnLSTM (w/o emb) & 0.23 ($\pm 0.06$)  & 0.15 ($\pm 0.07$) & 0.57 ($\pm 0.07$) \\
    \midrule
    Dummy & 0.11 ($\pm 0.08$)  & 0.14 ($\pm 0.06$)  & 0.10 ($\pm 0.04$)  \\ 
    \bottomrule
    \end{tabular}
    }
  \caption{Comparison of MLP and LSTM models for predicting hedges}
  \label{tab:overall_results}
\end{table*}

\subsection{Implementation Details}

In order to address the imbalance in our dataset, where the ratio of hedge to non-hedge instances is approximately 1:10, we used the Synthetic Minority Over-sampling Technique (SMOTE) \citep{chawla2002smote} for each model to augment our learning process. SMOTE is a popular method that generates synthetic examples in a dataset to counteract its imbalance. Given the variable nature of model performance, we implemented a 5-fold cross-validation strategy to evaluate the models. In order to account for the imbalanced nature of the dataset, we opted to use a lower number of folds in the cross-validation process. By choosing 5 folds instead of a higher number, we aimed to ensure that each fold would contain a sufficient representation of samples from each class. The model that delivered the best performance during this cross-validation process was then chosen to make predictions on the test set. For the neural models, we adjusted the loss function to account for class imbalance, thereby compelling the models to accommodate less frequent classes more effectively. The code is available in \url{https://github.com/neuromaancer/hedge_prediction}

\section{Results}

\subsection{Classification Results}

To answer the research question \ref{rq1}, we conducted classification experiments on different models. Table \ref{tab:overall_results}  offers an in-depth comparison of multiple machine learning models for predicting hedges in a peer-tutoring dataset.  We also incorporated a dummy classifier for comparison, which generates predictions in accordance with the class distribution observed in the training set. The performance metrics are F1 score, precision and recall, all of which include confidence intervals ($\alpha = 0.05$). The dataset is composed of several types of input features described in Section \ref{subsec:features}.  The models used different combinations of these inputs. (w/o emb) indicates that the model uses only the features without turn embeddings. If not specified, the model uses all features plus turn embeddings.

From Table \ref{tab:overall_results}, the LightGBM and XGBoost models without embeddings achieved relatively low scores for F1 scores, precision and recall, indicating limited performance in terms of balanced precision and recall. The MLP models, particularly those using only embeddings, showed a remarkable recall of 74\%, but at the cost of reduced precision. The LSTM model using only turn embeddings demonstrated balanced performance across all metrics, achieving the highest precision of 19\% and a competitive F1 score of 0.28. However, the attention-based LSTM (AttnLSTM) model did not significantly outperform the standard LSTM model in any metric.

The inclusion of turn embeddings significantly impacts model performance. Models with only embeddings perform better in terms of F1 score and recall, suggesting that the semantic information captured in these embeddings, which represented the semantic information of turns, is crucial for hedge prediction.  Second, models without embeddings also performed reasonably well in F1 score, implying that other features such as rapport, conversational strategies, tutoring strategies, nonverbal behaviors, and contextual information are also important. These features should not be overlooked.
\begin{table*}[!ht]
    \centering
    \resizebox{0.8\textwidth}{!}{%
        \begin{tabular}{l|ccccccc}
        \toprule
            \diagbox[]{\footnotesize{Model}}{\footnotesize{Feature}} & 
            \textbf{\footnotesize{\textit{N/A}}} & 
            \textbf{\footnotesize{\textit{Rapport}}} & \textbf{CS} & \textbf{\textit{TS}} & \textbf{\textit{NB}} & \textbf{\textit{ConInfo}} & \textbf{\textit{DialAct}} \\ 
            \midrule
            XGBoost & 0.24 ($\pm 0.07$) & 0.15 ($\pm 0.08$) & 0.10 ($\pm 0.08$) & 0.15 ($\pm 0.09$) & \textbf{0.08} ($\pm 0.07$) & 0.10 ($\pm 0.08$) & 0.12 ($\pm 0.08$) \\ 
            LightGBM & 0.24 ($\pm 0.07$) & 0.16 ($\pm 0.08$) & \textbf{0.09} ($\pm 0.08$) & 0.10 ($\pm 0.07$) & 0.10 ($\pm 0.10$) & 0.12 ($\pm 0.09$) & 0.13 ($\pm 0.08$) \\ 
            \midrule
            LSTM & 0.25 ($\pm 0.05$) & 0.24 ($\pm 0.05$) & 0.26 ($\pm 0.06$) & 0.24 ($\pm 0.06$) & 0.22 ($\pm 0.06$) & 0.25 ($\pm 0.07$) & \textbf{0.21} ($\pm 0.06$) \\ 
            AttnLSTM & 0.23 ($\pm 0.06$) & \textbf{0.20} ($\pm 0.06$) & 0.22 ($\pm 0.05$)  & 0.25 ($\pm 0.05$) & 0.24 ($\pm 0.05$) & 0.23 ($\pm 0.07$)  & 0.22 ($\pm 0.06$) \\ 
            MLP & 0.26 ($\pm 0.06$) & 0.25 ($\pm 0.06$)  & 0.25  ($\pm 0.06$) & 0.26 ($\pm 0.06$)  & 0.25 ($\pm 0.06$)  & 0.27 ($\pm 0.06$) & \textbf{0.21} ($\pm 0.07$)  \\ 
            \midrule
        \end{tabular}%
    }
    
    \caption{F1 scores after the feature ablation, \textit{CS}: Conversational Strategies; \textit{TS}: Tutoring Strategies; \textit{NB}: Nonverbal Behaviors; \textit{ConInfo}: Contextual Information; \textit{DialAct}: Dialogue Act.}
    \label{table:ablation}
\end{table*}

The LightGBM and XGBoost models, which only use features without turn embeddings, also display competitive performance compared to the MLP, LSTM, and AttnLSTM models using all features. This suggests that although turn embeddings provide valuable information for hedge prediction, models can still achieve satisfactory results even without them. The AttnLSTM models, which incorporate attention mechanisms, do not show significant improvements over the regular LSTM models. This could be due to the limited amount of data available, which cannot unleash the potential of the attention mechanism. 

Since good performance can also be achieved using the extracted features, in order to answer our research question 2, in the next subsections we will mainly investigate the importance of features in predicting hedges.

\subsection{Features Explanation with Shapley values}

Shapley values \citep{hart1989shapley}, originating from cooperative game theory, have emerged as a powerful model-agnostic tool to explain the predictions of machine learning models. This approach provides a way to fairly distribute the contribution of each feature to the overall prediction for a specific instance. By calculating the Shapley value for each feature, we gain insight into the importance of individual features within the context of a specific prediction. This interpretability technique has been adopted across various machine learning models. In this study, we use Shapley values to interpret the contributions of extracted features in our classification models using the SHAP python package \citep{NIPS2017_7062}.

Figure \ref{fig:shap} in the Appendix illustrates the importance of each feature for prediction when only features are used as input to different prediction models. The importance of features within the models can differ depending on their architectures. For simplicity, we identify the features that frequently appear in these 4 figures as significant indicators. Therefore, we have selected some of the most representative features in predicting hedges in Table \ref{tab:features_valence}.


\begin{table}[htbp]
  \centering
  \resizebox{0.7\linewidth}{!}{%
    \begin{tabular}{ll}
    \toprule
    \textbf{Features} & \textbf{Valence} \\
    \midrule
    correctness & \textcolor{red}{+} \\
    no gaze from tutor & \textcolor{blue}{-} \\
    problem id & \textcolor{blue}{-} \\
    rapport & \textcolor{blue}{-} \\
    tutee's deep question & \textcolor{blue}{-} \\
    tutee's gaze at tutor & \textcolor{blue}{-} \\
    tutee's pre-test & \textcolor{blue}{-} \\
    tutor's gaze at elsewhere & \textcolor{blue}{-} \\
    tutor's praise & \textcolor{blue}{-} \\
    \bottomrule
    \end{tabular}
    }
\caption{Features and their Valences}
  \label{tab:features_valence}
\end{table}

Based on Table \ref{tab:features_valence}, certain features have a significant impact on the likelihood of using hedges in tutoring conversations. Rapport has a negative valence, suggesting that higher rapport between the participants results in a lower likelihood of hedges being used. This confirms the finding cited above, that hedges are more frequent in low rapport interaction \citep{madaio2017using}. Interestingly, the ``problem ID'' feature also has a negative valence, indicating that as the complexity or difficulty of the problem increases, the likelihood of using hedges decreases. This could be because tutors tend to be more assertive or confident when addressing more challenging problems.

Moreover, certain conversational features such as ``tutee's deep question'' and ``tutor's praise'' have a negative valence, implying that these actions tend to decrease the likelihood of hedges. This could be because deeper questions or praise might indicate a more open and confident dialogue, thus reducing the need for hedges.

The table also reveals a negative correlation between various non-verbal cues such as ``no gaze from tutor'', ``tutee's gaze at tutor'', and ``tutor's gaze at elsewhere'', and the occurrence of hedges.  When the tutor is not gazing at the tutee, the likelihood of hedges decreases. The tutee's gaze at the tutor and the tutor's gaze at elsewhere are negatively associated with the use of hedges. This could indicate that when tutors' attention is focused elsewhere, they are attending less to how best to convey instruction or correction. To our knowledge, this is the first demonstration that specific nonverbal cues substantially influence the likelihood of a hedge being used in the succeeding turn of peer-tutoring interactions.

\subsection{Ablation Study}


We next examine the aforementioned models with different features ablated from input. This approach allows us to identify which features, when absent, lead to the best or worst performance in each model, implying that these features may not have contributed positively (or negatively) to the model's performance. Our study considered 6 groups of features: Conversational Strategies (\textit{CS}), Tutoring Strategies (\textit{TS}), Nonverbal Behaviors (\textit{NB}), Contextual Information (\textit{ConInfo}), Dialogue Act (DialAct), and Rapport. 

Table \ref{table:ablation} shows the different F1 scores as a consequence of removing the different features. For XGBoost and LightGBM, the worst performance is observed when \textit{NB} and \textit{CS} were removed, respectively, which implies that these features may provide important information for these models. The LSTM and MLP models showed a significant drop in performance when the \textit{DialAct} feature was removed, suggesting a substantial dependency of these models on the \textit{DialAct} feature for their prediction capabilities. Interestingly, the best performance of AttnLSTM was achieved when the rapport feature was removed, suggesting that the attention mechanism could compensate for loss of rapport. 

\section{Conclusion and Future Work}

This study presents an effective approach to predict where hedges occur in peer-tutoring interactions using classic ML models. Our results show the importance of considering various types of input features, such as turn embeddings, rapport, conversational strategies, tutoring strategies, nonverbal behaviors, and contextual information. Moreover, Shapley values applied to the predictions of the different models show, for the first time, that the gaze of both tutor and tutee may play a critical role in predicting hedges. This observation is substantiated by subsequent ablation studies, where classic classification models, like XGBoost and LightGBM, experienced a significant decline in F1 score when removing nonverbal behavior features.

For future work, several directions can be pursued. First, the investigation of hedge generation in the context of expert tutors could provide valuable insights into how experienced tutors use hedges differently and how these differences might affect learning outcomes. Second, incorporating reinforcement learning techniques to enhance specific aspects of the interaction, such as learning performance, could improve the practical applications of our findings.

\section*{Acknowledgments}
We thank the anonymous reviewers for their helpful feedback. We express sincere gratitude to the members of the ArticuLabo at Inria Paris for their invaluable assistance in the successful completion of this research, and to the members of the ArticuLab at Carnegie Mellon University in Pittsburgh for answering our questions about their prior work. This study received support from the French government, administered by the Agence Nationale de la Recherche, as part of the "Investissements d'avenir" program, with reference to ANR-19-P3IA-0001 (PRAIRIE 3IA Institute).



\bibliography{refs}
\bibliographystyle{acl_natbib}

\newpage

\onecolumn
\appendix

\begin{center}
\textbf{Appendix: SHAP Value Graphs}

The vertical axis indicates the mean contribution of the feature over the model decision. The horizontal
axis indicates how the distribution of features influences the model decision.

\end{center}

\begin{figure*}[htbp]
\centering
\begin{subfigure}{0.45\textwidth}
    \centering
    \includegraphics[width=\linewidth]{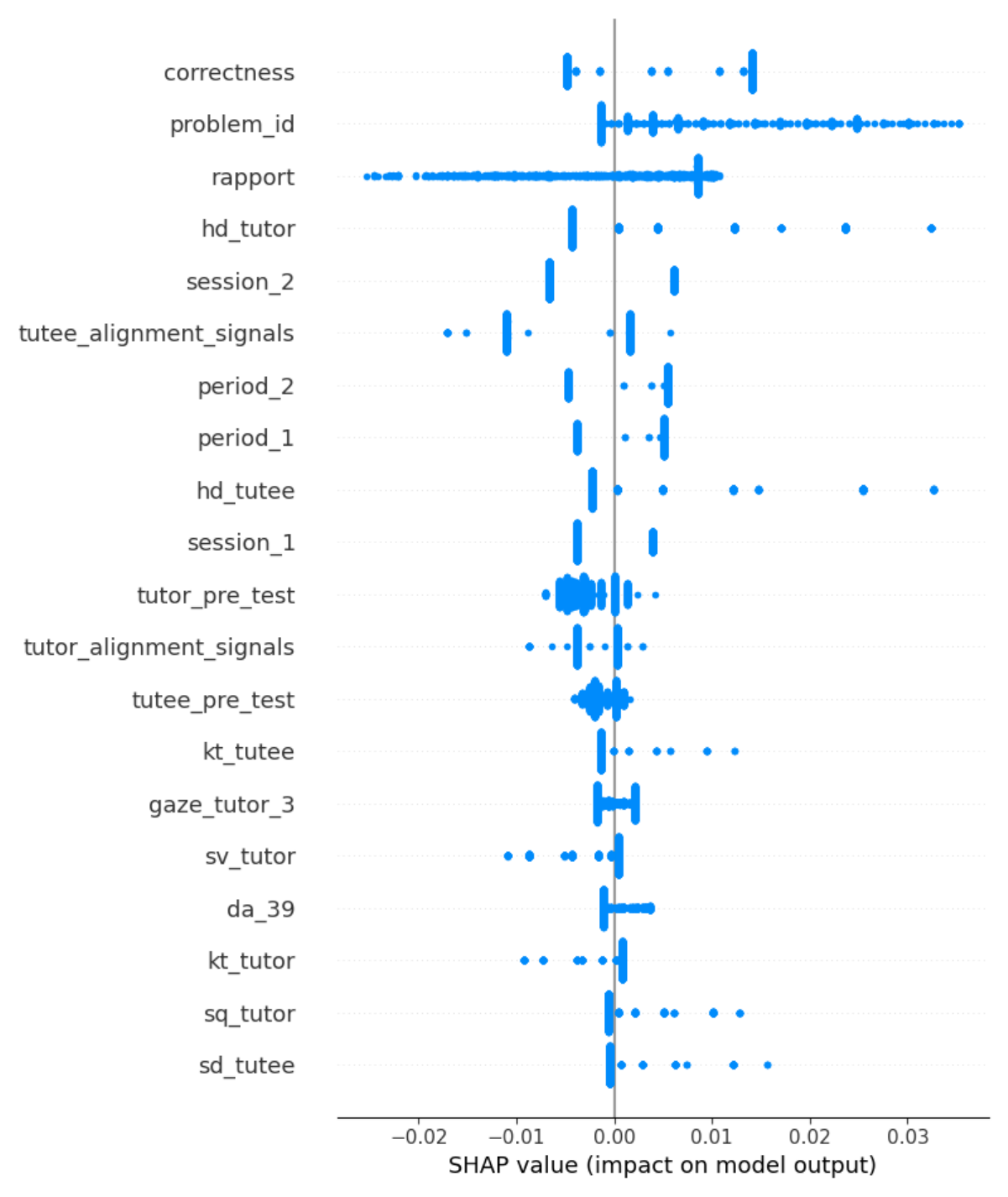}
    \caption{Feature Importance for AttnLSTM (without emb)}
    \label{fig:lstm_shap}
\end{subfigure}
\hfill
\begin{subfigure}{0.45\textwidth}
    \centering
    \includegraphics[width=\linewidth]{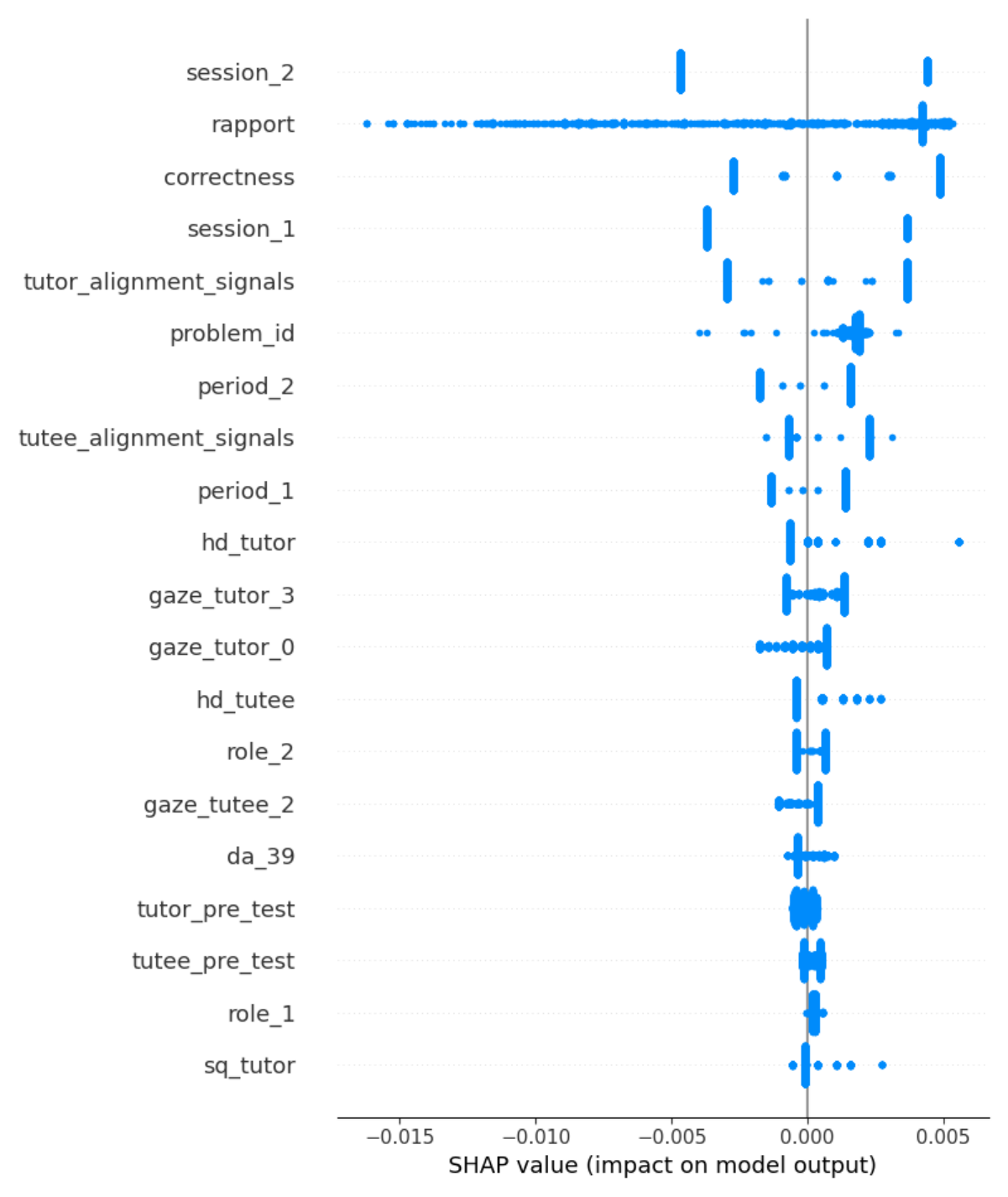}
    \caption{Feature Importance for MLP (without emb)}
    \label{fig:mlp_shap}
\end{subfigure}

\begin{subfigure}{0.45\textwidth}
    \centering
    \includegraphics[width=\linewidth]{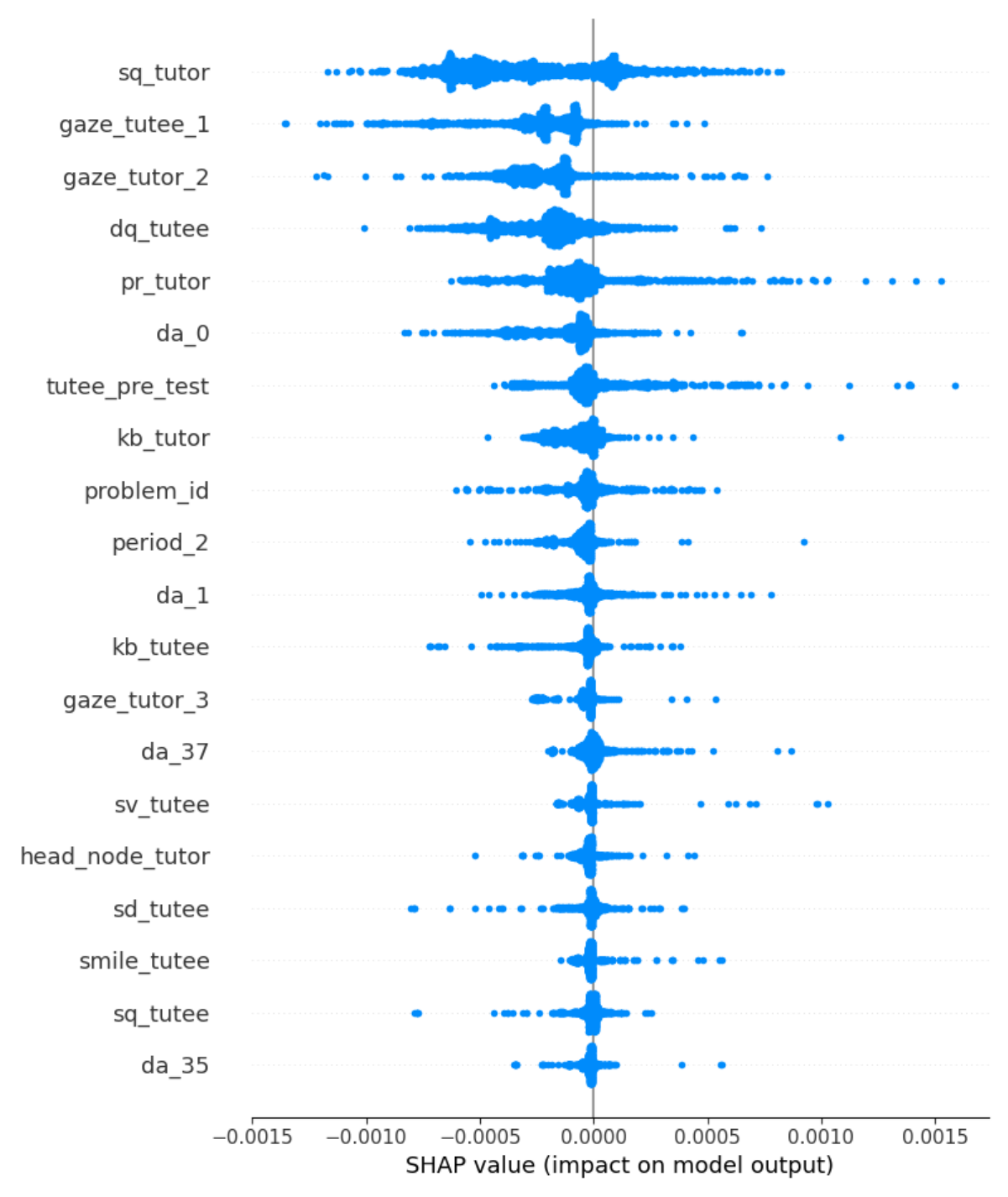}
    \caption{Feature Importance for XGBoost}
    \label{fig:xgboost_shap}
\end{subfigure}
\hfill
\begin{subfigure}{0.45\textwidth}
    \centering
    \includegraphics[width=\linewidth]{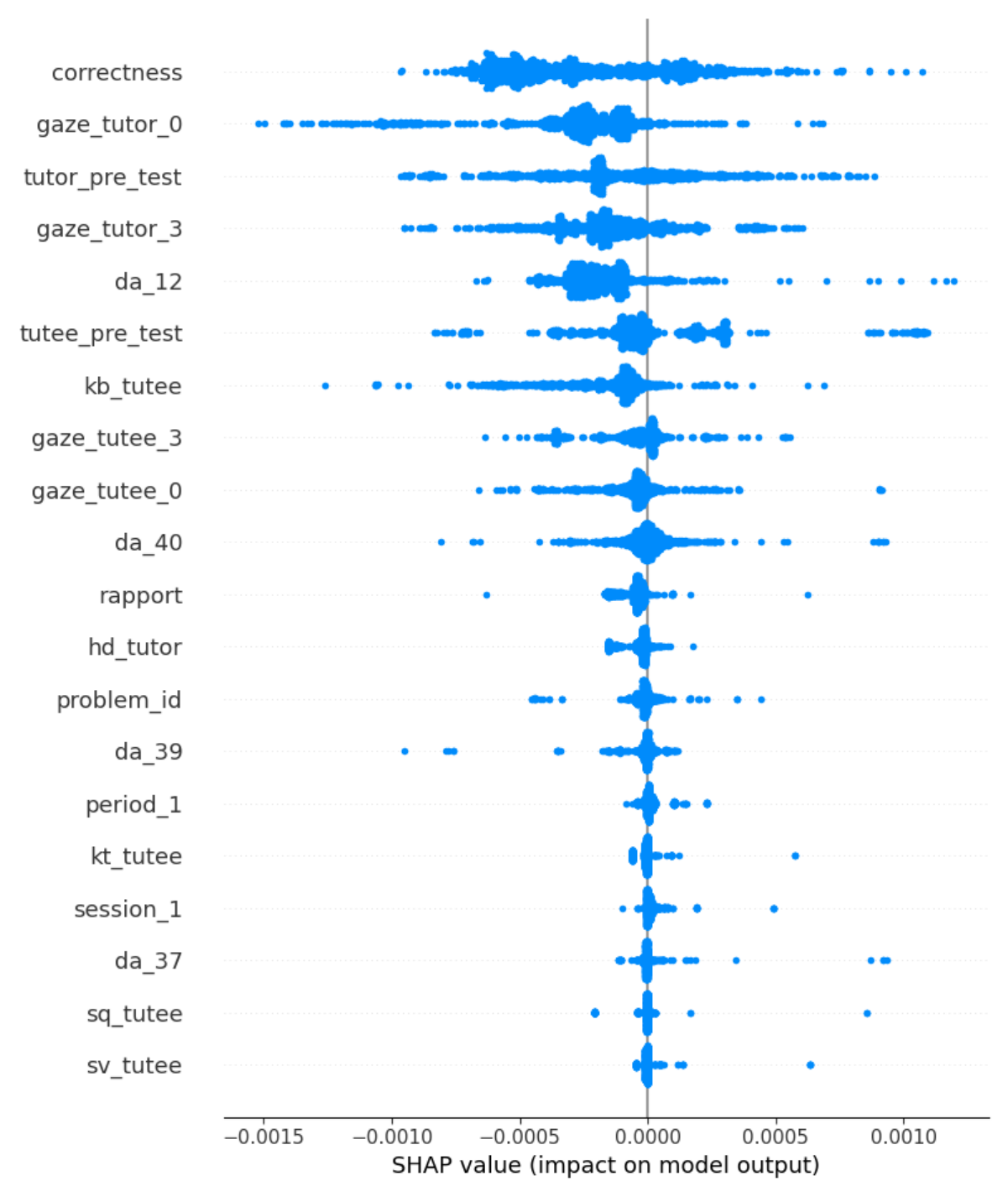}
    \caption{Feature Importance for LightGBM}
    \label{fig:lightgbm_shap}
\end{subfigure}
\caption{Feature Importance for Different Models}
\label{fig:shap}
\end{figure*}

\end{document}